\documentclass[conference, a4paper]{IEEEtran}
\usepackage{flushend,cuted}
\usepackage{stfloats}
\usepackage{multirow}

\usepackage{cite}

\ifCLASSINFOpdf
  \usepackage[pdftex]{graphicx}
  \graphicspath{{../pdf/}{../jpeg/}}
  \DeclareGraphicsExtensions{.pdf,.jpeg,.png}
\else
  \usepackage[dvips]{graphicx}
  \graphicspath{{../eps/}}
  \DeclareGraphicsExtensions{.eps}
\fi

\usepackage{amsmath}
\usepackage{amsfonts}
\usepackage{amssymb}

\usepackage{algorithmic}

\usepackage{array}

\usepackage{mdwmath}
\usepackage{mdwtab}

\usepackage{eqparbox}

\usepackage{caption}

\newcommand{\eg}{\emph{e.g.}}

\newcommand{\etal}{\emph{et al.}}

\begin{document}
%
\title{FaceCook: Face Generation Based on \\Linear Scaling Factors}


\author{\IEEEauthorblockN{Tianren Wang,
Can Peng,
Teng Zhang, 
Brian Lovell}
\IEEEauthorblockA{School of Information Technology and Electrical Engineering\\\
The University of Queesland,
Brisbane, Queensland 4067\\ Email: \{tianren.wang, can.peng, patrick.zhang\}@uq.edu.au, lovell@itee.uq.edu.au }
}

\maketitle

\begin{strip}
	\centering{\includegraphics[width=\textwidth]{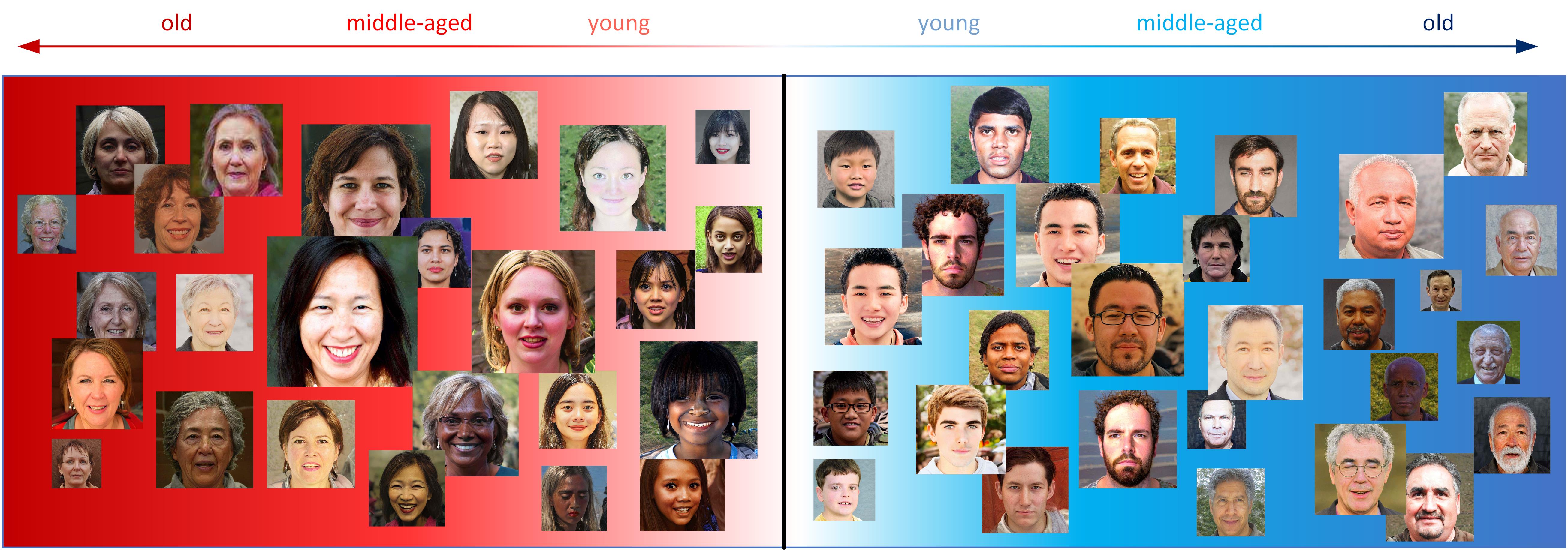}}
	\captionof{figure}{Several examples of synthesised images from our proposed method.
		Faces are arranged with respect to gender and age.
		Note that these images are directly generated from the latent vectors, rather than by editing randomly generated images.
	}
	\label{fig1}
\end{strip}

\begin{abstract}
	With the excellent disentanglement properties of state-of-the-art generative models, image editing has been the dominant approach to control the attributes of synthesised face images.
	However, these edited results often suffer from artifacts or incorrect feature rendering, especially when there is a large discrepancy between the image to be edited and the desired feature set.
	Therefore, we propose a new approach to mapping the latent vectors of the generative model to the scaling factors through solving a set of multivariate linear equations.
	The coefficients of the equations are the eigenvectors of the weight parameters of the pre-trained model, which form the basis of a hyper coordinate system.
	The qualitative and quantitative results both show that the proposed method outperforms the baseline in terms of image diversity.
	In addition, the method is much more time-efficient because you can obtain synthesised images with desirable features directly from the latent vectors, rather than the former process of editing randomly generated images requiring many processing steps.
\end{abstract}

\IEEEpeerreviewmaketitle

\section{Introduction}
Controllable semantic image editing enables an interactive approach to change the attribute of images with a few clicks, \eg\ age, gender, and other general features.
Classic approaches to this task use a Generative Adversarial Net (GAN)\cite{goodfellow2014generative} to learn a latent space and suitable latent-space transformations.
However, current approaches often suffer from attribute edits that are entangled, global image identity changes, and the occurrence of image artifacts.
To address these issues, a milestone work, named StyleGAN\cite{karras2019style,karras2020analyzing}, proposed to sample the style vectors of generated output from an intermediate latent space $\mathcal{W}$, rather than directly from the normal distribution $\mathcal{Z}$ as previous works had proposed.
With this work as the backbone, the performance of image editing improved greatly in terms of image quality, image diversity, and attribute accuracy.
However, the attribute accuracy of the edited outcome is highly dependent on the original image input.
As experimental results showed, the edited images tend to have poor attribute accuracy when the attribute combination of the original images is significantly different from that of the desired combination.
For example, this becomes a problem if users need to synthesise an ``old man" image from a ``young girl.
"  Although this issue can be alleviated to some extent by upscaling the factors which control the related attribute intensity, other unrelated attributes will change due to the generation model not being fully disentangled.
Besides, the overly large scaling factors tend to induce significant artifacts in the resulting images.

Therefore, we propose a new mechanism for the image editing task, or more precisely, due to the situation where current state-of-the-art models are still not fully disentangled, we propose to generate images from the semantic labels directly, rather than using the labels to guide the image editing of a randomly generated image.
The contributions of our work can be summarised as follow:

\begin{itemize} \item Different from the previous image editing method, we propose to generate images directly from the semantic labels to improve the attribute accuracy of the resulting image.
	\item To enhance the attribute disentanglement of the backbone, we propose to further project the style vectors of the intermediate latent space to a combination of scaling factors.
	      Inspired by \cite{shen2021closed}, we achieve this by solving the instance-level system of linear equations whose coefficient matrix is the eigenvectors of the mapping model of StyleGAN\cite{karras2020analyzing}.
	\item As shown in our experimental results, we achieve both higher attribute accuracy and greater image diversity compared to previous methods.
	      The work also shows that the generated images can have more accurate and richer attribute generation if equipped with a higher accuracy multi-attribute classifier.
\end{itemize}

The paper is organised as follows.
In the second section, we discuss relevant works whose topics include feature disentanglement and image editing.
In the third section, we explain our proposed method in detail.
Next, the quantitative and quantitative experimental results between our method and the baseline show the effectiveness and achievement of the proposed method.
In the final section, besides concluding our work, we highlight the limitations of the current work and then discuss the roadmap for the next stages of this research.

\section{Related Work}
\subsection{Feature Disentanglement}
Researchers take the degree of feature disentanglement as an important metric to evaluate their methods in computer vision tasks.  This is especially true for the image generation task because disentanglement facilitates the control of the attributes of the generated output.
To this end, Chen \etal\
\cite{chen2016infogan} first proposed to maximise the mutual information between the attribute classes and the random noise input used to generate samples.
However, due to the random noise still being sampled from the normal distribution like the original GAN\cite{goodfellow2014generative}, features cannot be well disentangled because the normal distribution assumptiondistorts the original distribution of the training data.
Later Liu \etal\
\cite{liu2018exploring} adopted an Autoencoder model as the backbone and further disentangled the latent vectors of training samples into identity-related and identity-irrelevant categories.
Specifically, the identity-irrelevant branch is implemented by adversarial training, where the branch tries to output the features which have the same identity entropy as the later Softmax classification.
Nonetheless, the visualisation of the features of the identity-irrelevant branch indicated that the extracted features are still relevant to the training samples to some extent.
This causes the distribution of the same semantic meaning features to be disjoint and scattered unevenly in feature space --- this makes it difficult to model features of different semantic meanings with known distributions.
After this work, a milestone work, named StyleGAN\cite{karras2019style,karras2020analyzing}, provided a feasible disengagement method that improved the benchmark of both image quality and diversity.
The main idea of the work to disentangle features is to map the random noise vectors sampled from the normal distribution $\mathcal{Z}$ to an intermediate latent space $\mathcal{W}$.
A major benefit of this architecture is that the latent space $\mathcal{W}$ does not have to support sampling according to any fixed distribution; its sampling density is induced by a learned piecewise continuous mapping \cite{karras2019style}.
\subsection{Image Editing}
Due to the state-of-the-art performance of StyleGAN\cite{karras2020analyzing} and promising theoretical support, many works are based on it, especially focusing on fine-grained image editing.
In StyleGAN\cite{karras2020analyzing}, users have no access to control the attributes generation outcome of the synthetic images.
Among these, Wang \etal\
\cite{wang2021faces} proposes to extract semantic label vectors from the text input, and then use the differentiated results between the semantic label vectors and the image classification vectors to guide the feature modification of the randomly generated images.
The experimental results show the method succeeds in generating images with desired features with good diversity, but the percentage of an image generation batch which correctly shows all of the desired features is low.
The reason for this is because the low accuracy of the multi-attribute classifier fails to recognise the features that the input images have.
Besides, although the work uses StyleGAN\cite{karras2020analyzing} as the backbone, the image editing is still conducted by modifying the random noise vectors sampled from the normal distribution, which means the work did not take any advantage of the disentanglement ability of StyleGAN\cite{karras2020analyzing}.
At the same time, there have been several works focusing on it.
Among these works, H\"ark\"onen \etal\
\cite{harkonen2020ganspace} proposes to identify important latent directions based on the Principal Components Analysis (PCA) of the latent space vectors.
To do so, they sampled over 10000 latent space vectors and then computed a PCA on these vectors, which gives a basis $\mathcal{V}$ for $\mathcal{W}$.
Given a new image defined by $\boldsymbol{w}$, it can be edited by varying a combination of PCA coordinates of the basis $\mathcal{V}$.
Shen \etal\
\cite{shen2021closed}  noted that this method has some defects.
First, the sampled vectors need to be huge in number or the vectors will not fully cover the multi-modal data distribution.
Second, the enormous number of sampling operations is time-consuming.
Therefore, they propose a closed-form factorization algorithm for latent semantic discovery by directly decomposing the pre-trained weights of the official StyleGAN\cite{karras2020analyzing} models.
Besides, compared to other methods, the method is more robust in finding the principle feature directions because it ameliorates the influence of the bias on the sampling data.

\section{Methodology}
In this section, we introduce our method via three subsections.
In our study, we focus on  two crucial semantic labels when describing human faces, age and gender, and generate images directly from these labels, rather than editing  randomly generated images.
To do this, we take advantage of the closed-form factorisation algorithm proposed by\cite{chen2016infogan} to project the latent vectors of our latent space into the scaling factors of the linear space and then use a pre-trained classifier which can output a continuous outcome to classify the randomly generated images according to their pre-defined semantic rules. This approach classifies the corresponding scaling factors at the same time with a different value range.
As a next step to achieve better diversity of the generated images, we propose to uniformly sample the scaling factors in each channel within the corresponding value range.
In the following subsections, we will elaborate on the methodology.
\subsection{Preliminaries}
\subsubsection{Assumption of Generation Mechanism}
Assume a generative model which can fully disentangle the feature render of the output.
In other words, given a random vector, the model can demodulate all the feature information, w.r.t feature classes and scaling factors.
We can derive: \begin{equation} G(\mathbf{z}) \Leftrightarrow G(a_1x_1 + a_2x_2 + .
	..+ a_nx_n) = G(\boldsymbol{A}\mathbf{x})
	\label{eqn_1}
\end{equation}
where \(\boldsymbol{A}=\{a_1, a_2,..., a_n\}\) defines a feature space that can be seen as a hyper coordinate system.
Similarly, \( \boldsymbol{x} \) is a vector defining the location of the generated sample.
Therefore, we can conclude that the motivation of image generation and image editing is identical --- the aim is to relocate the sample point along a certain feature direction with a certain distance.
Thus, the goal, for now, is to find the hyper coordinate system and the corresponding coordinate of each sample.

\subsection{Unsupervised Semantic Factorization}
Inspired by \cite{shen2021closed}, we also adopt the parameters of the pre-trained StyleGAN\cite{karras2020analyzing} model to compute the principal components of the semantic feature space.
To do this, we solve the following optimization problem \begin{equation} \mathbf{f}^* = \mathop{\arg\max}_{\mathbf{f}\in\mathbb{R}^\mathit{d}: \mathbf{f}^T\mathbf{f}=1}\\\|\boldsymbol{W}\mathbf{f}\|^2_2 \label{eqn_2} \end{equation} where \(\\\| \cdot \|_2\) is $l_2$ norm and $\boldsymbol{W}$ is a matrix consisted of the pre-trained model weight parameters.
This problem aims at finding the directions that can cause large variations after the projection of $\boldsymbol{W}$.
In practice, to find the $k$ most significant directions $\bf{N}=\{\bf{n}_1,\bf{n}_2,.
	..,\bf{n}_k\}$, we introduce the Lagrange multipliers $\{\lambda_i\}^k_{i=1}$ into Eq(2) as:
\begin{equation}
	\begin{aligned}
		\mathbf{F}^* & =\mathop{\arg\max}_{\mathbf{F}\in\mathbb{R}^{d \times k}}\sum_{i=1}^k \|\boldsymbol{W}\mathbf{f}\|^2_2-\sum_{i=1}^k \lambda_i(\mathbf{f}^T\mathbf{f}-1)                   \\
		             & =\mathop{\arg\max}_{\mathbf{F}\in\mathbb{R}^{d \times k}}\sum_{i=1}^k(\mathbf{f}_i^T\boldsymbol{W}^T\boldsymbol{W}\mathbf{f}_i-\lambda_i\mathbf{f}^T\mathbf{f}+\lambda_i)
	\end{aligned}
	\label{eqn_3}
\end{equation}
By taking the partial derivative on each $\mathbf{n}_i$, we obtain \begin{equation} 2\boldsymbol{W}^T\boldsymbol{W}\mathbf{f}_i-2\lambda_i\mathbf{f}_i=0 \label{eqn_4} \end{equation} We can see this is actually also a principal component analysis (PCA) process.
But different from\cite{harkonen2020ganspace}, the PCA components are computed from the weight parameters rather that the sampled vectors.

\subsection{Scaling Factors Resampling}
With the above two steps, We have obtained the feature coordinate system consisted of the $k$ most dominant feature directions.
Given $\mathbf{N}$ randomly sampled vectors, we can project them to the linear scaling factor space by: \begin{equation} \mathbf{N}=\boldsymbol{W\alpha} \label{eqn_5} \end{equation} where $\mathbf{\alpha}$ is a set of coordinates of each sample.
According the pre-defined rules, the coordinate set can be divided into subsets which have $c$ classes $\mathbf{\alpha}=\{\alpha^1,\alpha^2,.
	..,\alpha^c\}$.
For each subset, we extract the minimum and maximum respectively.
Then we can get a coordinate value range for each class:

\begin{equation} \{\boldsymbol{\alpha^1},\boldsymbol{\alpha^2},.
	..,\boldsymbol{\alpha^c\}}\Rightarrow
	\left(
	\begin{array}{ccc}
			[\alpha^1_{min},\alpha^1_{max}]            \\
			\left[\alpha^2_{min},\alpha^2_{max}\right] \\
			\cdots                                     \\
			\left[\alpha^c_{min},\alpha^c_{max}\right]
		\end{array}
	\right)
	\label{eqn_6}
\end{equation}

Then we uniformly sample the new factor combinations from the target feature value range, and then compute the new latent vector through the hyper coordinate system $\boldsymbol{W}$:

\begin{equation} \begin{aligned} \boldsymbol{\alpha^i}=Uni([\alpha^i_{min} & ,\alpha^i_{max}])\quad(i=1,2,.
		..,c)                                                                                            \\
		                                          & \boldsymbol{w^*}=\boldsymbol{W}\boldsymbol{\alpha^i}
	\end{aligned}
	\label{eqn_7}
\end{equation}

The reason for this is illustrated in Figure 2.
On the left side, the figure shows a possible scaling factor distribution in a certain feature direction which is obtained from the random noise vectors.
We can see the scaling factors is unevenly distributed in the value range.
This is because even though the model can disentangle the feature to some extent, the samples, sampled from the normal distributions in most circumstances, still couple to the normal distribution.
While on the right side, given a fixed value range, we propose to sample the scaling factors from the uniform distribution.
In this way, we  obtain some factor combinations which have a low probability of occurrence or may not ever occur in the original sample method.

\begin{figure}[!htb]
	\centering
	\includegraphics[width=3.5in]{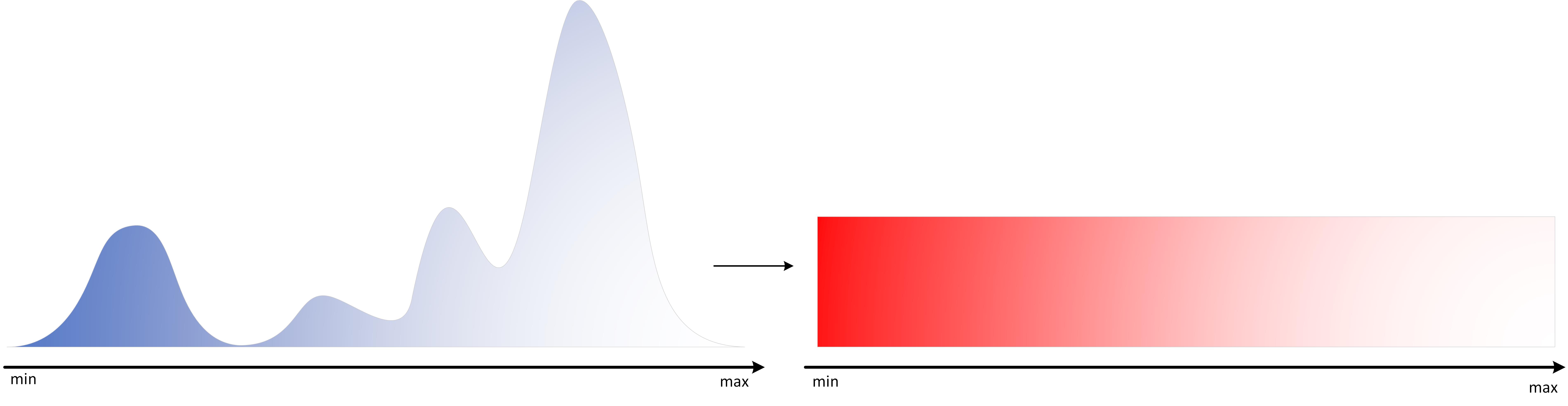}
	\caption{Visulisation of possible data distribution before and after the resampling operation.}
	\label{fig_2}
\end{figure}
In addition, the goal of our study is to control the feature generation by manipulating the factors in the high-dimensional space.
As shown in Figure 3, the samples with the same semantic labels can be enclosed with an irregular curved surface body (transparent sphere).
The sample points scatter unevenly within each body.
While for our proposed sampling method, the irregular curved bodies change to cubes (Figure 4).
Although the cubes may enclose some sample points which do not belong to the corresponding semantic labels, the sampled points distribute evenly in the corresponding semantic label subspace, which means that this method can make more use of the disentangled hyper coordinate system.

\begin{figure}[!htb]
	\centering
	\includegraphics[width=3.5in]{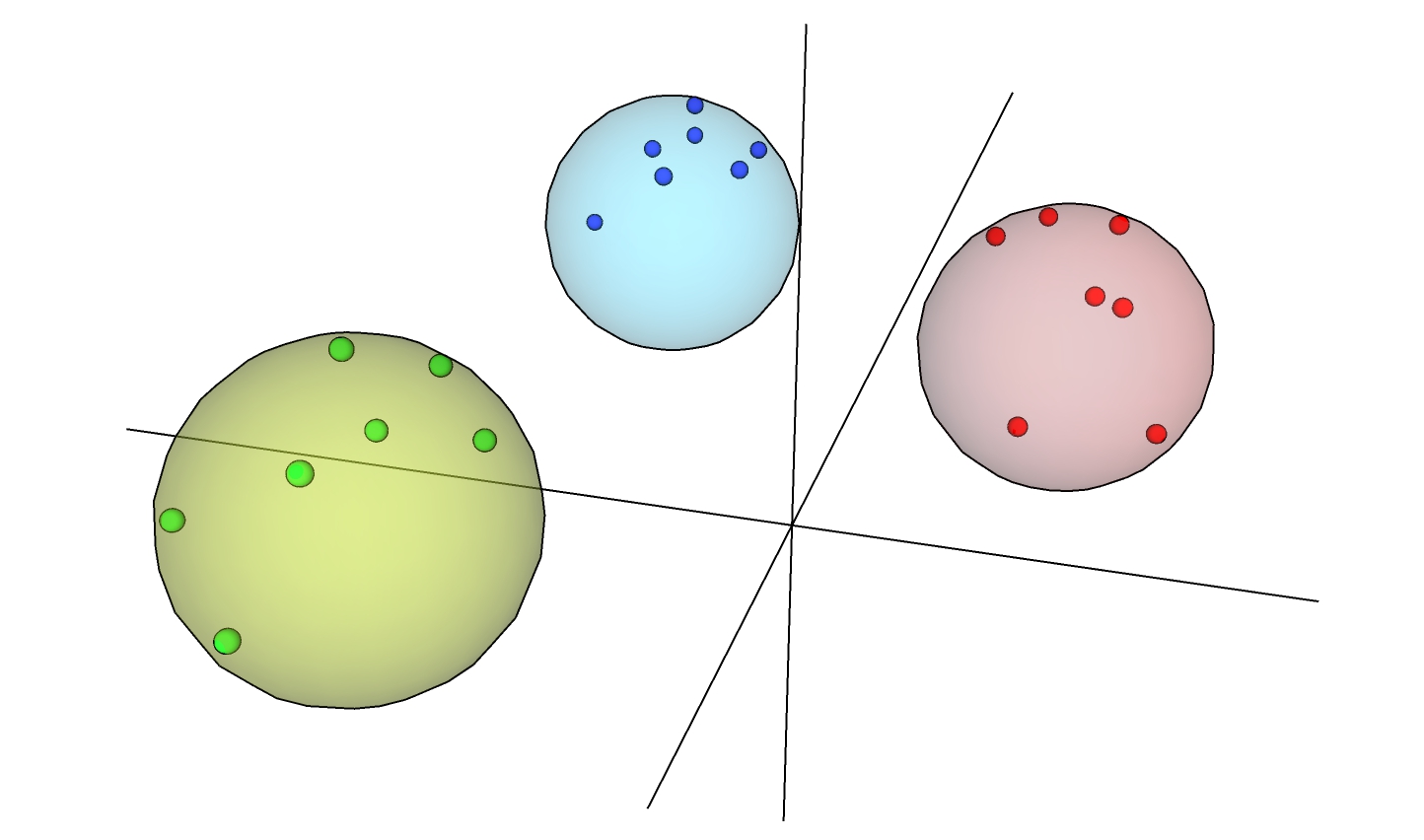}
	\caption{Visualisation of latent vectors of different semantic classes in the hyper coordinate system of the StyleGAN\cite{karras2020analyzing} model.}
	\label{fig_3}
\end{figure}

\begin{figure}[!htb]
	\centering
	\includegraphics[width=3.5in]{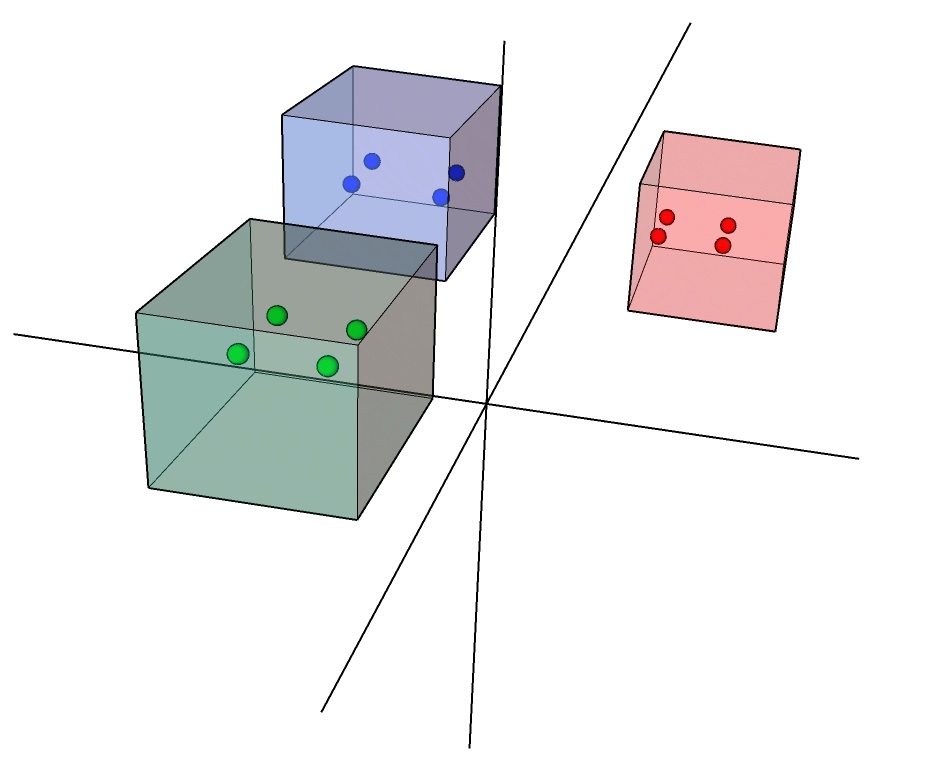}
	\caption{Visualisation of latent vectors of different semantic classes of uniform sampling in the hyper coordinate system.}
	\label{fig_4}
\end{figure}

\section{Experiments \& Evaluation}
\subsection{Implementation}
We use the models of StyleGAN\cite{karras2020analyzing} as our backbone.
For the semantic label classifier, we choose the DEX\cite{rothe2015dex} to classify the gender and age of the generated images.
There are several reasons for only to classify the gender and age aspects.
Firstly, these two features are the most common and important ones when describing faces.
Besides, as far as we know, there does not exist a classifier that can classify other fine-grained features of faces with high accuracy.
Although there are some models\cite{dong2018imbalanced,zhuang2018multi,mao2020deep} which are trained with CelebA\cite{liu2015faceattributes} dataset whose samples are labelled with 40 classes, the data imbalance issue and model architecture limits addressing multi-label classification.
Therefore, in this study, we only focus on controlling the age and gender rendering of the generated images.
As mentioned in Section 3, we can obtain different value ranges by the rules.
In practice, we label the age feature with "young", "middle-aged", and "old" semantic labels.
Therefore, we have six value ranges to sample from arranged as follow:

\begin{table}[!
		t]
	\renewcommand{\arraystretch}{1.5}
	\small
	\caption{Semantic Labels and Corresponding Value Ranges}
	\label{table_1}
	\centering
	\begin{tabular}{cccc}
		\hline
		Gender & Young(\textless{30})              & Middle-aged(30-60)                & Old(\textgreater{60})             \\
		\hline
		Female & $[\alpha^1_{min},\alpha^1_{max}]$ & $[\alpha^2_{min},\alpha^2_{max}]$ & $[\alpha^3_{min},\alpha^3_{max}]$ \\
		Male   & $[\alpha^4_{min},\alpha^4_{max}]$ & $[\alpha^5_{min},\alpha^5_{max}]$ & $[\alpha^6_{min},\alpha^6_{max}]$ \\
		\hline
	\end{tabular}
\end{table}

We can uniformly sample as the number of factor combinations as desired from a certain value range.
These factor vectors are transferred to latent vectors and then images eventually.
\subsection{Qualitative Evaluation}
Figure 5 shows some experimental results of our method.
We use the same random seeds under different semantic value range.
We can see the images shows good performance in terms of ageing and gender transformation.
Some features can change reasonably, like the grey hair and glasses for elder people, while other irrelevant features keep unchanged, like background and hair colour for young and middle-aged people.

\begin{figure}[!htb]
	\centering
	\includegraphics[width=3.5in]{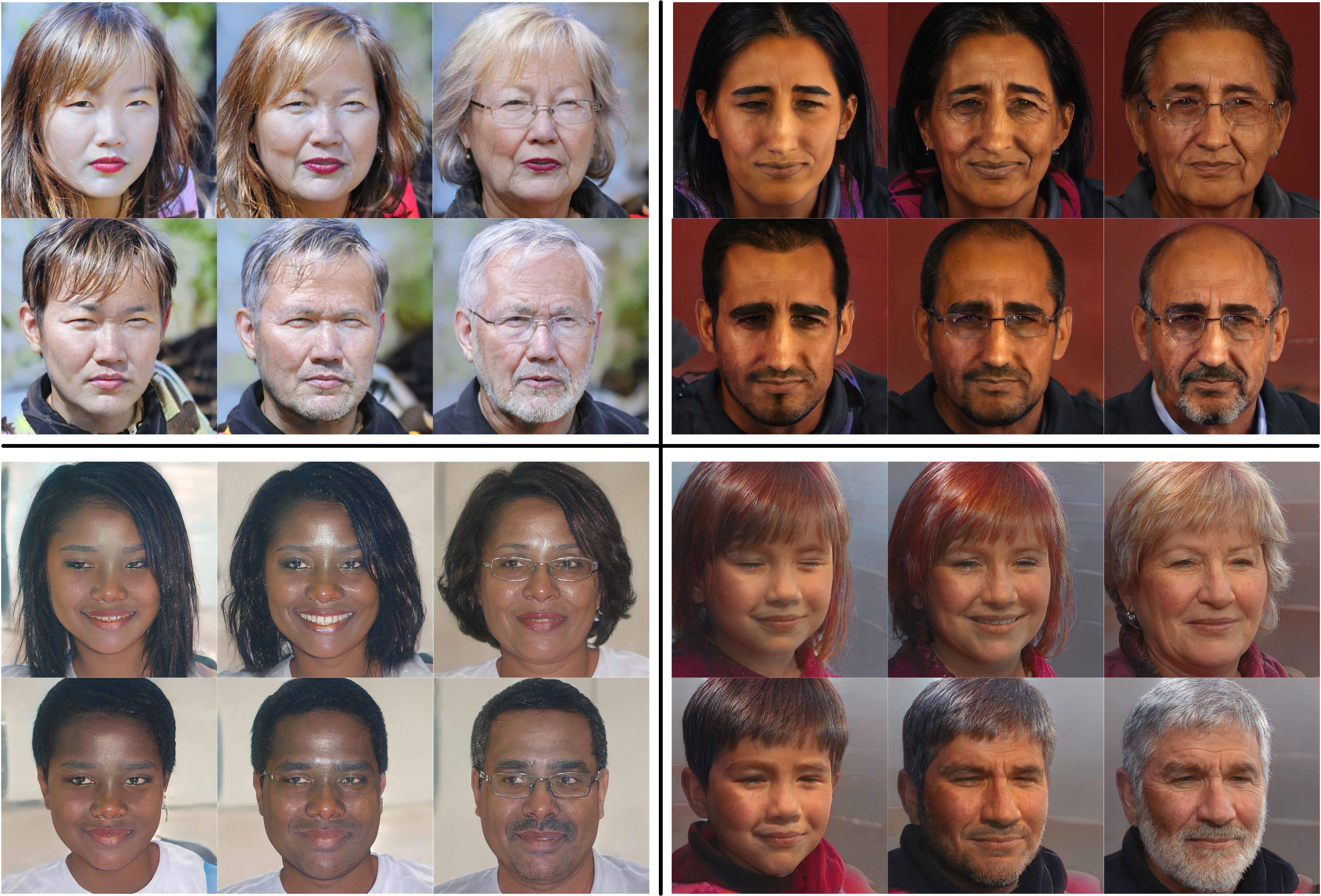}
	\caption{Some experimental results of the proposed method.
		We fix the random seeds when sampling factors under different semantic label value ranges.
		From the figure, we can see the age and gender features change naturally with inducing artifacts.
		At the same time, the proposed method well preserved other irrelevant features like identification.
	}

	\label{fig_5}
\end{figure}

\begin{table}[!htb]
	\renewcommand{\arraystretch}{1.5}
	\scriptsize
	\caption{Evaluation on LPIPS Metrics}
	\label{tab:table2}
	\centering
	\begin{tabular}{l|llll}
		\hline
		\multirow{2}{*}{Method} & \multicolumn{2}{l|}{Ours} & \multicolumn{2}{l}{StyleGAN\cite{karras2020analyzing}}                                         \\ \cline{2-5}
		                        & Female                    & \multicolumn{1}{l|}{Male}                              & Female            & Male              \\
		\hline
		Young                   & $\bf{.6814^{\pm.002}}$    & $\bf{.7184^{\pm.002}}$                                 & $.6388^{\pm.002}$ & $.6716^{\pm.002}$ \\
		Middle-aged             & $\bf{.6786^{\pm.002}}$    & $\bf{.6984^{\pm.002}}$                                 & $.6330^{\pm.002}$ & $.6574^{\pm.002}$ \\
		Old                     & $\bf{.6473^{\pm.002}}$    & $\bf{.6684^{\pm.002}}$                                 & $.6321^{\pm.002}$ & $.6569^{\pm.002}$ \\
		\hline
	\end{tabular}
\end{table}

\begin{table}[!t]
	\renewcommand{\arraystretch}{1.5}
	\normalsize
	\caption{User Study on Feature Accuracy}
	\label{tab:table3}
	\centering
	\begin{tabular}{cccc}
		\hline
		Gender & Young  & Middle-aged & Old    \\
		\hline
		Female & 97.0\% & 92.5\%      & 88.5\% \\
		Male   & 85\%   & 90\%        & 85\%   \\
		\hline
	\end{tabular}
\end{table}

\subsection{Quantitative Evaluation}
We evaluate the results of our method on the aspects of image diversity and feature accuracy.
For the first aspect, we use the LPIPS\cite{zhang2018unreasonable} metric to compare the image diversity of the images generated with our method and the images generated from the StyleGAN\cite{karras2020analyzing} model.
In practice, 1000 images are generated and sampled from each semantic class.
For the output of StyleGAN\cite{karras2020analyzing} model, we also use the pre-trained DEX\cite{rothe2015dex} classifier to recognise the age and gender of the generated images until 1000 images are collected for each class.

The values in Table \ref{tab:table2} indicate that the proposed method outperforms the original StyleGAN\cite{karras2020analyzing} with respect to image diversity.
The higher the scores, the better the diversity of the images.
Because the uniformly sampled points scatter more evenly in the hyper coordinate system, the average distance between sample points is larger than that of the sample points of StyleGAN\cite{karras2020analyzing}.

Apart from image diversity, we also investigate the feature accuracy of the generated images of our method.
We synthesise 200 images in each class and perform do a user study.
Table \ref{tab:table3} shows the consequences of whether the user thinks the images belong to their categories.
We can see the ``young female" class has the highest accuracy.
The reasons behind this are various.
First, the dataset used to train StyleGAN\cite{karras2020analyzing} is imbalanced in each class.
From the CelebA\cite{liu2015faceattributes} Database notes we deduce that this dataset has a larger percentage of ``young female" images, leading to the subspace of ``young female" in the hyper coordinate system being larger than others.

\section{Conclusion}
In this study, we propose to map the latent vectors of StyleGAN\cite{karras2020analyzing} to the scaling factor combinations through solving multivariate linear equations.
The coefficient of the equations is the eigenvectors of the parameters of the pre-trained model of StyleGAN\cite{karras2020analyzing}, which consist of a hyper coordinate system.
The qualitative and quantitative results show that the proposed method outperforms the baseline StyleGAN in terms of image diversity.
In addition, it is much more time-efficient because you can obtain the synthesised images with your desirable features directly from the latent vectors, rather than editing randomly generated images with many complex steps.

The study still has some limitations which are worth diving into in future research.
Firstly, only two features is nor really sufficient to describe faces.
We will focus on how to control a much larger range of features of different semantic labels in the later research.
In addition, we find that although the images generated by the proposed method have higher diversity, they lose some global scale feature variation as well, such as the variation of face angles.
One assumption of thsi work is that the subspace of a certain semantic label may not exist only in one place.
Therefore, for one semantic label, we may lose some data points located in other parts of the hyper coordinate system.

	{\small
		\bibliographystyle{IEEEtran}
		\bibliography{dicta_tianren}
	}
\end{document}